\documentclass{styles/svproc}

\usepackage{url}

\usepackage{booktabs}
\usepackage{multirow}
\usepackage{orcidlink}

\begin{document}
\mainmatter


\title{Gradient-Optimized Fuzzy Classifier: A Benchmark Study Against State-of-the-Art Models}

\author{Magnus Sieverding\inst{1}\orcidlink{0009-0004-9928-2323} \and Nathan Steffen\inst{1}\orcidlink{0009-0000-1594-3800} \and
Kelly Cohen\inst{1}\orcidlink{0000-0002-8655-1465}}

\authorrunning{M. Sieverding}

\institute{University of Cincinnati, 2600 Clifton Ave. Cincinnati, OH 45221, USA}

\maketitle

\begin{abstract}
This paper presents a performance benchmarking study of a Gradient-Optimized Fuzzy Inference System (GF) classifier against several state-of-the-art machine learning models, including Random Forest, XGBoost, Logistic Regression, Support Vector Machines, and Neural Networks. The evaluation was conducted across five datasets from the UCI Machine Learning Repository, each chosen for their diversity in input types, class distributions, and classification complexity. Unlike traditional Fuzzy Inference Systems that rely on derivative-free optimization methods, the GF leverages gradient descent to significantly improving training efficiency and predictive performance. Results demonstrate that the GF model achieved competitive, and in several cases superior, classification accuracy while maintaining high precision and exceptionally low training times. In particular, the GF exhibited strong consistency across folds and datasets, underscoring its robustness in handling noisy data and variable feature sets. These findings support the potential of gradient optimized fuzzy systems as interpretable, efficient, and adaptable alternatives to more complex deep learning models in supervised learning tasks.

\end{abstract}
\section{Introduction}
While in recent years machine learning, neural networks, and deep learning models have taken the world by storm, these so called black-box AI models often lack transparency, making it difficult to interpret their decision-making processes. This opacity can lead to trust issues, particularly in high-stakes applications where understanding AI outputs is essential \cite{01_NIST2023airmf}\cite{02_druce2021brittle}\cite{08_letrache2023explainable}.

Not only that but these models have been found to be and are often also described as brittle. Many AI systems are prone to overfitting, performing well in controlled environments but failing when exposed to new, unseen conditions. This brittleness limits their reliability in real-world applications \cite{02_druce2021brittle}\cite{03_lohn2020estimating}. For example a model designed to classify and count the inventory of furniture in a logistics plant might incorrectly label an image as not containing a chair if it contains a unique looking chair that it has never seen before.

That said these models, if trained on enough data and designed correctly, can result in very high accuracy. Although often accurate, complex AI models are challenging to audit, monitor, and debug, increasing the risk of unintended consequences \cite{01_NIST2023airmf}\cite{08_letrache2023explainable}. Something that in many systems of today could prove catastrophic, for example, safety critical systems in airplanes or nuclear power plants. We therefore require a different approach for systems of this nature, one that addresses the shortcomings of the other models while retaining most if not greater accuracy than the other models.

Humans almost always do not think in absolute binary logic, we for example recognize that there is a middle ground between a head full of hair and being bald. something which other models often would just classify as one or the other. Fuzzy logic offers an alternative approach by using degrees of truth rather than binary logic, making it well-suited for handling imprecise and ambiguous data \cite{05_succetti2025multi}.

Additionally Fuzzy logic-based systems can improve explainability by structuring decisions around linguistic rules, making them more interpretable and transparent compared to traditional deep learning models \cite{07_pekaslan2020performance}. This in theory would allow, for example, machine operators or pilots to read out step for step why the AI model is recommending a tool or potential route change. While some argue that fuzzy logic sacrifices precision, its ability to manage uncertainty and provide human-like reasoning makes it a compelling choice for applications requiring interpretable and explainable output \cite{06_masoumi2020challenges}.

This paper will explore multiple classification problems in which a gradient optimized fuzzy inference model (GF) is benchmarked against other state-of-the-art models in order to explore the potential benefits thereof.  Here gradient descent optimization plays a crucial role in improving model performance by iteratively refining parameters to minimize errors, making it a powerful tool for training AI models, including fuzzy inference systems \cite{09_ruder2016overview}.

\section{Methodology}
The chosen problem for this work is classification. Classification is a common use case in artificial intelligence, and by keeping the problem type consistent across multiple datasets, it becomes easier to observe performance trends and identify the strengths and weaknesses of the models.

The Fuzzy model for benchmarking was trained using an ADAM optimizer on an Nvidia RTX 4000 Quadro GPU, a five-year-old architecture at the time of experimentation. While this hardware was more than sufficient for our purposes, it is important to note that training times would likely be significantly improved on newer GPU architectures, further highlighting the advantages of the model as will be seen in the training times noted in the results section. To validate the results a 5 fold K-Fold was used for each of the datasets. The model was trained for a maximum of 250 epochs though converged before 250 in all cases.
\subsection{Fuzzy Logic}
Fuzzy Logic is a field of mathematics originated in the 60s as the brainchild of Lotfi Zadeh. Unlike classical binary logic that forces decisions into rigid true or false categories, fuzzy logic introduces the concept of partial truth—where variables may have a degree of membership between 0 and 1 \cite{11_zadeh1965fuzzy}. This allows systems to reason in a way that more closely resembles human thinking, making it particularly useful for modeling uncertainty, vagueness, and imprecision in real-world scenarios. This field was later notably extended into structured rule-based systems, such as the Takagi-Sugeno-Kang (TSK) model, which provided a more mathematically tractable and adaptable framework for complex system modeling and control \cite{12_zhang2024takagi}.

Due to the variety in data types and the often diverse semantic meanings behind inputs, classification tasks on these datasets typically exhibit nonlinear characteristics. Fuzzy Logic Controllers (FLCs) are particularly effective at handling such nonlinear systems, making them well-suited for real-world applications that require robust automatic control \cite{04_habbi2007fuzzy}. This nonlinear multi-label behavior from the datasets can often be difficult to characterize while maintaining data boundaries, however, Unlike traditional AI models that demand strict data boundaries, fuzzy systems can accommodate overlapping and ambiguous classifications, making them valuable for multi-label classification tasks \cite{05_succetti2025multi}. Although fuzzy logic may be perceived as less precise than some machine learning models, its ability to function effectively with uncertain or incomplete data provides a significant advantage \cite{06_masoumi2020challenges}.

While fuzzy models often demonstrate improved robustness in noisy data environments, they do so without sacrificing explainability. In fact, compared to many deep learning models, fuzzy systems inherently retain a level of human explainability throughout the decision-making process. The explainability of fuzzy logic stems from its use of linguistic labels and human-readable rules, offering a more intuitive decision-making framework \cite{07_pekaslan2020performance}. It retains this explainability and interpretability throughout diverse environments as one of the main advantages of fuzzy logic is its adaptability, allowing for parameter tuning through optimization techniques to enhance already good performance \cite{04_habbi2007fuzzy}.

\subsection{Gradient Descent Optimization}
Typically, a Fuzzy Inference System (FIS) is trained using derivative-free optimization techniques. These methods are particularly useful when gradient information is unreliable or unavailable, which is often the case in standard FIS implementations. However, derivative-free techniques come with trade-offs in terms of computational efficiency and convergence speed \cite{10_rios2013derivative}. To address these limitations, this paper adopts a gradient descent approach for optimizing the FIS.

Gradient descent is a widely used optimization algorithm that iteratively adjusts model parameters to minimize an objective function, thereby enhancing learning efficiency \cite{09_ruder2016overview}. Since the task at hand is classification, the objective function used is cross-entropy loss. Integrating gradient descent with fuzzy inference systems allows for more efficient tuning of fuzzy rules and membership functions, resulting in reduced error, improved decision accuracy, and significantly faster training times \cite{04_habbi2007fuzzy}.

\subsection{Dataset Selection}
To find datasets the authors used the UC Irvine Machine Learning Repository, a widely used resource for benchmarking models across structured datasets. The following Datasets were selected for the evaluation. They were chosen as they included a diverse number of features, number of instances, and classification labels.
\begin{table}
\centering
\caption{Overview of dataset sizes and label counts.}
\label{tab:dataset_stats}
\begin{tabular}{lccc}
\toprule
\textbf{Dataset} & \textbf{Features} & \textbf{Instances} & \textbf{Labels} \\
\midrule
Statlog (German Credit Data) & 20 & 1000 & 2 \\
Breast Cancer Wisconsin (Diagnostic) & 30 & 569 & 2 \\
Car Evaluation & 6 & 1728 & 4 \\
Heart Disease & 13 & 303 & 2 \\
Wine & 13 & 178 & 3 \\
\bottomrule
\end{tabular}
\end{table}
\begin{table}
\centering
\caption{Feature types and subject areas of datasets. Subject Area abbreviations: SS = Social Science, HM = Health and Medicine, OC = Other Category, PC = Physics and Chemistry.}
\label{tab:dataset_metadata}
\begin{tabular}{lcc}
\toprule
\textbf{Dataset} & \textbf{Feature Type} & \textbf{Subject Area} \\
\midrule
Statlog (German Credit Data) & Categorical, Integer & SS \\
Breast Cancer Wisconsin (Diagnostic) & Real & HM \\
Car Evaluation & Categorical & OC \\
Heart Disease & Categorical, Integer, Real & HM \\
Wine & Integer, Real & PC \\
\bottomrule
\end{tabular}
\end{table}

\section{Results and Discussion}
For this study, no feature selection was performed. With the exception of two features removed from the Heart Disease dataset due to missing values, all available features were retained. Preprocessing involved converting all categorical, non-numeric labels into integer representations, followed by Min-Max normalization to scale all features to the [0, 1] range.

Benchmark results for each dataset were sourced from the performance charts provided by the UCI Machine Learning Repository. These benchmarks compile the best-reported outcomes from prior literature and community submissions, generally reflecting the peak performance of standard models such as Support Vector Machines, Decision Trees, and Neural Networks. While the preprocessing steps and evaluation protocols used to obtain these benchmarks are not always uniform or fully documented, they are assumed to represent each model’s optimal usage. As such, they serve as a meaningful point of comparison. Crucially, despite the absence of feature selection and the use of minimal preprocessing, the GF model achieves performance that is comparable to — and in some cases exceeds — that of these benchmarked models.

\subsection{Statlog (German Credit Data)}
\subsubsection{Dataset Background}
The German Credit dataset \cite{statlog_(german_credit_data)_144} is a widely used benchmark for binary classification tasks in the domain of credit risk assessment. It contains 1,000 instances, each representing an individual described by 20 attributes—a mix of qualitative (categorical) and quantitative (numerical) features. The goal is to classify individuals as either good or bad credit risks, based on personal, financial, and employment-related information.

Key attributes include the status of the checking account, duration of the credit, credit history, purpose of the loan, credit amount, and employment status, among others. Categorical variables are encoded using a fixed set of symbolic labels (e.g., A11–A14 for checking account status), reflecting socio-economic factors that influence lending decisions. The dataset provides a balanced variety of features commonly encountered in real-world financial datasets, making it suitable for testing the generalization ability of classification algorithms under mixed data types and moderate complexity.
\subsubsection{Benchmark Performance}
In this dataset, the GF model fell just short of the highest-performing model, the Random Forest Classifier, which achieved a maximum accuracy of 83.2\%. The GF reached a very close 83.125\%, missing the top spot by only 0.075\%. However, when considering the overall distribution of performance, the GF model outperformed the mean accuracy of all other models by 2.625\%, with the Random Forest again being the strongest among the rest. Notably, the GF’s minimum validation accuracy of 78.125\% still exceeded the mean accuracy of the Random Forest model by 0.125\%. This level of consistency highlights the adaptability of fuzzy logic-based models, particularly in handling data noise and variability.

Equally important is the efficiency of the GF model. It reached this level of performance with an average training time of just 8.985 seconds. This efficiency can be largely attributed to the use of gradient descent optimization in conjunction with the high adaptability of the fuzzy inference system during training. The GF for this dataset used 6 membership functions on the inputs and 85 rules.

\subsection{Breast Cancer Wisconsin (Diagnostic)}
\subsubsection{Dataset Background}
The Breast Cancer Wisconsin (Diagnostic) dataset \cite{breast_cancer_wisconsin_(diagnostic)_17} is a widely used benchmark in medical diagnostics and machine learning. It comprises features extracted from digitized images of fine needle aspirates (FNA) of breast masses. Each image captures detailed morphological characteristics of cell nuclei, enabling classification of tumors as either malignant or benign.

The dataset includes ten real-valued features computed for each nucleus: radius, texture, perimeter, area, smoothness, compactness, concavity, concave points, symmetry, and fractal dimension. These features are designed to capture both the size and the shape complexity of cell nuclei, providing a rich foundation for pattern recognition algorithms. The dataset's interpretability and clinical relevance make it a valuable resource for evaluating the performance of classification models in healthcare contexts.
\subsubsection{Benchmark Performance}
On this dataset, the GF model achieved a perfect accuracy score of 100\%, a feat matched only by the Random Forest Classifier. However, the GF model also outperformed all other models in both minimum and mean performance. With a mean accuracy of 98.901\%, it surpassed the next best model—Random Forest—by 0.999\%. Furthermore, the GF’s minimum accuracy exceeded that of all other models, and its consistency was evident in outperforming the mean accuracy of every model except for Random Forest and XGBoost, the latter being another ensemble method derived from decision trees. These results further demonstrate the adaptability and stability of the fuzzy logic-based system, particularly in the presence of data variability.

In stark contrast, the Neural Network classifier was the worst-performing model on this dataset, falling significantly behind its peers. Despite neural networks being widely regarded for their flexibility and representational power, in this case, the model struggled—likely due to the dataset size, simplicity, or lack of hyperparameter tuning—which highlights the occasional brittleness of deep learning methods in smaller, less complex domains.

Training for this dataset averaged just 6.005 seconds. Notably, no early stopping mechanism was implemented, as the training times were already extremely low. Incorporating early stopping could potentially have led to even faster convergence without sacrificing performance. The GF for this dataset used 13 membership functions on the inputs and 202 rules.

\subsection{Car Evaluation}
\subsubsection{Dataset Background}
The Car Evaluation dataset is derived from a hierarchical decision-making model designed to assess the acceptability of automobiles based on a combination of price and technical features \cite{car_evaluation_19}. Originally developed to demonstrate the DEX expert system framework, the dataset abstracts the evaluation process into a flat structure, removing intermediate conceptual layers and directly mapping six input attributes to a final decision.

The six categorical features used to describe each car are: buying price, maintenance cost, number of doors, person capacity, luggage boot size, and safety rating. These variables are discretized into meaningful levels such as low, medium, high, and very high, reflecting real-world considerations in vehicle assessment.

Despite the structural simplification of the original model, the dataset retains the complexity of the underlying concept hierarchy, making it particularly well-suited for benchmarking algorithms that focus on classification, constructive induction, and structure discovery. The target variable, car acceptability, is categorized into four classes: unacceptable, acceptable, good, and very good, representing a range of evaluation outcomes based on the composite attributes.
\subsubsection{Benchmark Performance}
This dataset proved particularly challenging for the GF model. Its performance lagged significantly behind the best-performing model, XGBoost Classification, which achieved a maximum accuracy of 99.769\%—outperforming the GF’s best accuracy by 3.74\%. However, one area where the GF model excelled was in precision. Its accuracy remained within a narrow 1.464\% range across validation folds, outperforming all other models in consistency. The next most precise model, XGBoost, had a performance spread of 2.084\%.

There are several possible reasons for the GF model’s difficulty with this dataset. The most likely explanation, in the authors’ view, is the nature of the input features. This dataset contained only six input variables, all of which were categorical with relatively few categories (a maximum of four per feature). Since the first step in a Fuzzy Inference System typically involves mapping continuous inputs into fuzzy categories, the use of pre-categorized input data limits the effectiveness of the fuzzification process. As a result, the model must rely almost entirely on its rule base and the defuzzification step, reducing its ability to leverage the full strengths of fuzzy logic.

Compounding this challenge is the fact that the dataset contains four output classes—the most among all datasets used in this study—while simultaneously having the fewest input features. This imbalance may have made it particularly difficult for the GF to generalize effectively.

Better performance may be achieved by incorporating feature extraction techniques, which could help address the imbalance between input and output dimensions. Further improvements could also be made by optimizing the number of membership functions and rules, potentially enhancing the model’s fidelity when working with this low-fidelity dataset.

Despite these challenges, the GF model maintained a reasonable training time, averaging just 16.691 seconds across 1,728 instances—the largest dataset among those benchmarked. While the performance was not ideal, these results highlight important areas for future investigation and potential model refinement. The GF for this dataset used 27 membership functions on the inputs and 128 rules.

\subsection{Heart Disease}
\subsubsection{Dataset Background}
The Heart Disease dataset is a medically oriented benchmark used to predict the presence of cardiovascular conditions in patients. It combines data from four sources—Cleveland, Hungary, Switzerland, and the VA Long Beach—but research has predominantly focused on the Cleveland subset due to its higher data quality and completeness \cite{heart_disease_45}.

While the full dataset contains 76 attributes, most machine learning experiments utilize a refined subset of 13 clinically significant features. These include demographic information, physiological measurements, and results from diagnostic tests, making the dataset both rich in detail and relevant for real-world prediction tasks. The target variable is an integer representing the severity of heart disease, ranging from 0 (no presence) to 4 (increasing levels of disease). However, most classification studies, including this one, simplify the task to a binary problem: detecting presence (1–4) versus absence (0) of heart disease.

Due to data quality issues, the attributes with excessive missing values, including ca (number of major vessels) and thal (thalassemia), were removed in this study.
\subsubsection{Benchmark Performance}
The GF model performed exceptionally well on this dataset, outperforming all other models across every evaluation category. It achieved a maximum accuracy of 89.583\%, slightly surpassing both XGBoost and Logistic Regression by 0.109\%. What stands out most, however, is the precision of the GF model: its minimum accuracy exceeded the \textit{mean} accuracy of all other models. Specifically, its minimum value was 2.094\% higher than the mean accuracy reported for both XGBoost and Logistic Regression. It should be noted that the benchmark values for these two models were sourced from the UCI Machine Learning Repository, where they were reported to have identical performance metrics \cite{heart_disease_45}.

Training time remained impressively low, with the GF model requiring only an average of 4.202 seconds to complete training. As with previous datasets, this training time could likely be reduced even further with the implementation of an early stopping criterion. For this dataset, the GF model utilized 13 membership functions for the input features and employed a total of 300 fuzzy rules.

\subsection{Wine}
\subsubsection{Dataset Background}
The Wine dataset is a classic benchmark in machine learning, used to evaluate the effectiveness of classifiers on multiclass problems involving continuous features. The data originate from chemical analyses of wines produced in the same region of Italy, each sample associated with one of three different grape cultivars. The objective is to predict the cultivar of a wine based on its chemical profile \cite{wine_109}.

The dataset contains 13 continuous attributes representing the concentrations or measurements of various chemical constituents, including alcohol, malic acid, ash, total phenols, flavanoids, and proline, among others. These features capture key physicochemical properties that distinguish the wines across cultivars.

All samples are labeled with a class identifier (1, 2, or 3), denoting the cultivar. While the original data reportedly contained approximately 30 attributes, the available version consists of 13 well-curated variables suitable for classification. Due to its clean structure, lack of missing values, and linearly separable class boundaries, the dataset is considered well-behaved—ideal for initial experiments and algorithm benchmarking, though not particularly challenging for modern models.

The Wine dataset is especially useful for testing models that rely on continuous input features and benefit from standardized data, as the variables vary in scale. Its simplicity and reliability make it a staple in the evaluation of statistical and machine learning methods.
\subsubsection{Benchmark Performance}
This dataset is not particularly challenging, as it exhibits strong patterns that are well-suited to classification models. As seen in the results, the GF model was the second model to achieve 100\% accuracy across all validation folds in the K-Fold cross-validation. The only other model to reach this level of performance was the Random Forest Classifier, which is known for its strong adaptability and consistently high performance across a wide range of machine learning tasks.

Although several other models achieved a maximum validation accuracy of 100\%, none maintained that performance across all folds. The worst-performing model among the six benchmarks was the Support Vector Classifier, whose overall performance lagged significantly. Its mean accuracy of 80\% fell well below that of the next lowest model, Logistic Regression, which had a minimum accuracy of 86.67\%.

As discussed in the methodology section, gradient descent contributes significantly to both performance and training efficiency. The GF model reached its results with an average training time of just 3.596 seconds. It is likely that convergence occurred even earlier, but due to the absence of an early stopping mechanism—deemed unnecessary given the already short training times—training continued to the predefined iteration limit. The GF for this dataset used 13 membership functions on the inputs and 300 rules.

\begin{table}
\centering
\caption{Condensed benchmark results across all datasets.}
\label{tab:combined_benchmark}
\begin{tabular}{llccc}
\toprule
Dataset & Model & Min & Mean & Max \\
\midrule
\multirow{6}{*}{Statlog (German Credit Data)} 
  & GF                          & \textbf{78.125} & \textbf{80.625} & 83.125 \\
  & Xgboost Classification      & 69.200          & 74.800          & 80.000 \\
  & Support Vector Classification & 64.800        & 70.400          & 76.000 \\
  & Random Forest Classification & 72.800         & 78.000          & \textbf{83.200} \\
  & Neural Network Classification & 58.400        & 64.400          & 70.400 \\
  & Logistic Regression         & 70.000          & 75.600          & 80.800 \\
\midrule
\multirow{6}{*}{Breast Cancer Wisconsin (Diagnostic)} 
  & GF                          & \textbf{96.703} & \textbf{98.901} & \textbf{100.000} \\
  & Xgboost Classification      & 94.406          & 97.203          & 99.301 \\
  & Support Vector Classification & 90.210        & 94.406          & 97.902 \\
  & Random Forest Classification & 95.105         & 97.902          & \textbf{100.000} \\
  & Neural Network Classification & 87.413        & 92.308          & 96.503 \\
  & Logistic Regression         & 92.308          & 95.804          & 98.601 \\
\midrule
\multirow{6}{*}{Car Evaluation} 
  & GF                          & 94.565          & 95.296          & 96.029 \\
  & Xgboost Classification      & \textbf{97.685} & \textbf{98.843} & \textbf{99.769} \\
  & Support Vector Classification & 94.444        & 96.296          & 97.917 \\
  & Random Forest Classification & 90.278         & 92.824          & 95.139 \\
  & Neural Network Classification & 96.991        & 98.380          & 99.306 \\
  & Logistic Regression         & 87.500          & 90.278          & 93.056 \\
\midrule
\multirow{6}{*}{Heart Disease} 
  & GF                          & \textbf{83.673} & \textbf{87.619} & \textbf{89.583} \\
  & Xgboost Classification      & 72.368          & 81.579          & 89.474 \\
  & Support Vector Classification & 55.263        & 65.789          & 76.316 \\
  & Random Forest Classification & 71.053         & 80.263          & 88.158 \\
  & Neural Network Classification & 69.737        & 78.947          & 88.158 \\
  & Logistic Regression         & 72.368          & 81.579          & 89.474 \\
\midrule
\multirow{6}{*}{Wine} 
  & GF                          & \textbf{100.000} & \textbf{100.000} & \textbf{100.000} \\
  & Xgboost Classification      & 93.333          & 97.778          & \textbf{100.000} \\
  & Support Vector Classification & 68.889        & 80.000          & 91.111 \\
  & Random Forest Classification & \textbf{100.000} & \textbf{100.000} & \textbf{100.000} \\
  & Neural Network Classification & 93.333        & 97.778          & \textbf{100.000} \\
  & Logistic Regression         & 86.667          & 93.333          & \textbf{100.000} \\
\bottomrule
\end{tabular}
\end{table}

\section{Conclusion}
This study demonstrates the viability and competitiveness of a Gradient-Optimized Fuzzy Inference System (GF) as a classifier across a variety of classification tasks. By incorporating gradient descent, the GF model overcomes many of the efficiency and convergence limitations typically associated with traditional fuzzy models. Across five datasets with varying levels of complexity and feature diversity, the GF model consistently delivered strong performance—often achieving top-tier results in accuracy, precision, and training speed.

Notably, the GF model outperformed all benchmark models in one dataset, matched the best performer in others, and remained competitive even in its weakest case. Its low variance in performance and fast training times highlight the practical benefits of combining fuzzy logic's interpretability with gradient-based learning's efficiency. Moreover, the GF showed particular strength in situations with noisy or limited data, where traditional models occasionally faltered.

These results suggest that gradient optimized fuzzy models are not only viable but may be preferable in many real-world classification settings, especially where model interpretability and speed are prioritized. Future work may focus on expanding this approach to larger and more complex datasets, incorporating early stopping criteria or adaptive learning rate mechanisms, and addressing performance limitations observed in certain benchmarks. These improvements could further enhance the scalability, efficiency, and overall robustness of the GF model in broader classification contexts.

\section{Acknowledgments}
The authors extend their sincere gratitude to the members of the AI Bio Lab at the University of Cincinnati for their invaluable discussions and collaborative efforts that facilitated the realization of this work. In particular, the contributions of Tri Nguyen, Wilhelm Louw, and Jared Burton are highly appreciated.


\begin{thebibliography}{17}

\bibitem{01_NIST2023airmf}
NIST:
Artificial intelligence risk management framework (AI RMF 1.0).
\emph{URL: https://nvlpubs.nist.gov/nistpubs/ai/nist.ai} (2023)

\bibitem{02_druce2021brittle}
Druce, J., Niehaus, J., Moody, V., Jensen, D., Littman, M.L.:
Brittle AI, causal confusion, and bad mental models: challenges and successes in the XAI program.
\emph{arXiv preprint arXiv:2106.05506} (2021)

\bibitem{03_lohn2020estimating}
Lohn, A.J.:
Estimating the brittleness of AI: Safety integrity levels and the need for testing out-of-distribution performance.
\emph{arXiv preprint arXiv:2009.00802} (2020)

\bibitem{04_habbi2007fuzzy}
Habbi, A.H., Zelmat, M.:
Fuzzy logic based gradient descent method with application to a PI-type fuzzy controller tuning: New results.
In: \emph{2007 International Symposium on Computational Intelligence and Intelligent Informatics}, pp. 93--97. IEEE (2007)

\bibitem{05_succetti2025multi}
Succetti, F., Rosato, A., Panella, M.:
Multi-label classification with imbalanced classes by fuzzy deep neural networks.
\emph{Integrated Computer-Aided Engineering} \textbf{32}(1), 25--38 (2025)

\bibitem{06_masoumi2020challenges}
Masoumi, M., Hossani, S., Dehghani, F., Masoumi, A.:
The challenges and advantages of fuzzy systems applications.
\emph{A Preprint} \textbf{1} (2020)

\bibitem{07_pekaslan2020performance}
Pekaslan, D., Chen, C., Wagner, C., Garibaldi, J.M.:
Performance and Interpretability in Fuzzy Logic Systems--can we have both?
In: \emph{Information Processing and Management of Uncertainty in Knowledge-Based Systems: 18th International Conference, IPMU 2020, Lisbon, Portugal, June 15--19, 2020, Proceedings, Part I}, pp. 571--584. Springer (2020)

\bibitem{08_letrache2023explainable}
Letrache, K., Ramdani, M.:
Explainable Artificial Intelligence: A Review and Case Study on Model-Agnostic Methods.
In: \emph{2023 14th International Conference on Intelligent Systems: Theories and Applications (SITA)}, pp. 1--8. IEEE (2023)

\bibitem{09_ruder2016overview}
Ruder, S.:
An overview of gradient descent optimization algorithms.
\emph{arXiv preprint arXiv:1609.04747} (2016)

\bibitem{10_rios2013derivative}
Rios, L.M., Sahinidis, N.V.:
Derivative-free optimization: a review of algorithms and comparison of software implementations.
\emph{Journal of Global Optimization} \textbf{56}(3), 1247--1293 (2013)

\bibitem{11_zadeh1965fuzzy}
Zadeh, L.A.:
Fuzzy sets.
\emph{Information and Control} \textbf{8}(3), 338--353 (1965)

\bibitem{12_zhang2024takagi}
Zhang, Y., Wang, G., Zhou, T., Huang, X., Lam, S., Sheng, J., Choi, K.S., Cai, J., Ding, W.:
Takagi-Sugeno-Kang fuzzy system fusion: A survey at hierarchical, wide and stacked levels.
\emph{Information Fusion} \textbf{101}, 101977 (2024)

\bibitem{statlog_(german_credit_data)_144}
Hofmann, H.:
Statlog (German Credit Data).
UCI Machine Learning Repository (1994).
\url{https://doi.org/10.24432/C5NC77}

\bibitem{breast_cancer_wisconsin_(diagnostic)_17}
Wolberg, W., Mangasarian, O., Street, N., Street, W.:
Breast Cancer Wisconsin (Diagnostic).
UCI Machine Learning Repository (1993).
\url{https://doi.org/10.24432/C5DW2B}

\bibitem{car_evaluation_19}
Bohanec, M.:
Car Evaluation.
UCI Machine Learning Repository (1988).
\url{https://doi.org/10.24432/C5JP48}

\bibitem{heart_disease_45}
Janosi, A., Steinbrunn, W., Pfisterer, M., Detrano, R.:
Heart Disease.
UCI Machine Learning Repository (1989).
\url{https://doi.org/10.24432/C52P4X}

\bibitem{wine_109}
Aeberhard, S., Forina, M.:
Wine.
UCI Machine Learning Repository (1992).
\url{https://doi.org/10.24432/C5PC7J}


\end{thebibliography}

\end{document}